\PassOptionsToPackage{table}{xcolor}
\documentclass[11pt]{article}

\usepackage[preprint]{acl}
\usepackage{times}
\usepackage{latexsym}
\usepackage[T1]{fontenc}
\usepackage[utf8]{inputenc}
\usepackage{microtype}
\usepackage{inconsolata}
\usepackage{graphicx}
\usepackage{amsmath,amssymb,amsthm}
\allowdisplaybreaks
\usepackage{booktabs}
\usepackage{array}
\usepackage{multirow}
\usepackage{xcolor}
\usepackage{pifont}
\usepackage{float}
\usepackage{placeins}
\usepackage{enumitem}
\usepackage{algorithm}
\usepackage[noend]{algpseudocode}

\newtheorem{proposition}{Proposition}

\definecolor{ourrow}{RGB}{252,232,240}
\definecolor{gtrow}{RGB}{236,242,252}
\newcommand{\ours}{\textsc{Stone}}
\newcommand{\gt}{\ding{51}}
\newcommand{\nogt}{\ding{55}}
\newcommand{\conc}{$^{\dagger}$}

\title{Simplify, Agree, Distill:\\ Label-Free Self-Distillation with Stepping Stones}

\author{
  Dingyang Chen\textsuperscript{$\spadesuit$} \quad Jinfeng Xiao\textsuperscript{$\spadesuit$} \quad Zhenyu Zhang\textsuperscript{$\spadesuit$} \quad William Hu\textsuperscript{$\spadesuit$} \quad Ziyu Wang\textsuperscript{$\spadesuit$} \quad Yang Liu\textsuperscript{$\spadesuit$} \\
  \bfseries Shirley Zhang\textsuperscript{$\spadesuit$} \quad  Yuan Ling\textsuperscript{$\spadesuit$} \quad Xinyang Shen\textsuperscript{$\spadesuit$} \quad Qi Zhang\textsuperscript{$\clubsuit$} \quad Peijie Qiu\textsuperscript{$\spadesuit$} \\[4pt]
  \textmd{\textsuperscript{$\spadesuit$}Amazon \qquad \textsuperscript{$\clubsuit$}Worcester Polytechnic Institute} \\[2pt]
  \normalsize\textmd{\texttt{\{dingerwa,jfx,zhenyuzh,hthaoqi,ziywan,yngliun,congzhan,yualing,xinyans,peijieq\}@amazon.com}} \\
  \normalsize\textmd{\texttt{qzhang9@wpi.edu}}
}

\begin{document}
\maketitle

\begin{abstract}
Can a language model teach itself to reason without labels? On-policy self-distillation (OPSD) uses
the model as its own teacher, but strengthens that teacher with a ground-truth solution. The natural
label-free substitute, voting over the model's attempts, fails precisely on hard problems: attempts
disagree, and their most common answer is often wrong. We propose \ours{}, a label-free alternative
that moves agreement to easier related problems. The model writes these stepping stones, keeps those it solves consistently, and
shows the teacher their worked solutions, with final answers removed; the student sees only the
original problem. No step reads a label, and inference needs no extra context or model calls. Across
five math benchmarks, \ours{} achieves the highest average accuracy for Qwen3-4B and Qwen3-8B in
thinking mode. Without thinking mode, it gains 13.2 and 22.7 points, 2.5 and 2.7 times the gains of
OPSD with ground-truth solutions.
\end{abstract}

\section{Introduction}

\begin{figure}[!t]
  \centering
  \includegraphics[width=\columnwidth]{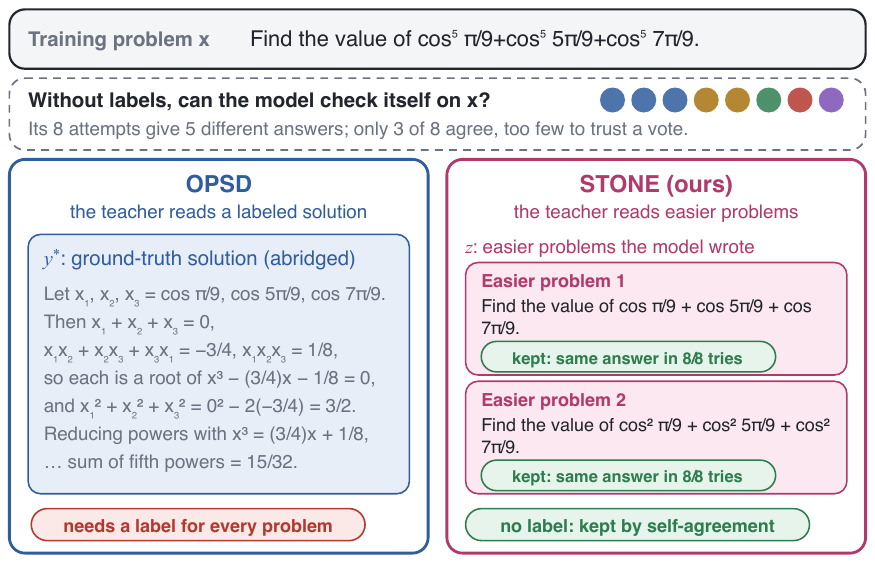}\\[4pt]
  \includegraphics[width=\columnwidth]{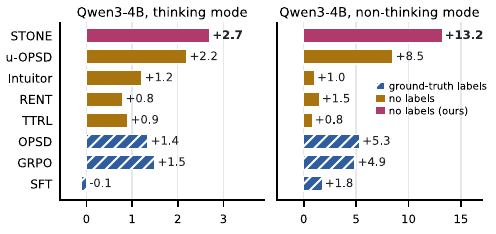}
  \caption{\textbf{A model can teach itself without labels.} \emph{Top:} the model's attempts disagree, so a
  vote cannot replace a label. OPSD shows the teacher a ground-truth solution; \ours{} shows easier
  problems the model wrote and answers consistently, with worked solutions. \emph{Bottom:} gain in average accuracy (points) over
  the base model for Qwen3-4B, in thinking and non-thinking modes. Table~\ref{tab:main} gives the
  results.}
  \label{fig:teaser}
\end{figure}

\emph{On-policy self-distillation} (OPSD) lets a language model learn from itself, but its teacher
still needs an answer key \citep{zhao2026self}. The student samples an attempt; the same model,
conditioned on a ground-truth solution, scores each token; and the student learns to match those
scores without seeing the solution. OPSD therefore removes the solution from the student, not from
the training data. Can the model construct useful teacher context on its own?

The natural label-free substitute is to trust the most common answer among the model's own attempts.
Yet voting fails precisely on the problems worth learning from. As Section~\ref{sec:diagnosis}
shows, among the 9{,}218 training problems on which Qwen3-4B is unsure, 28\% have no answer shared
by half of its attempts, and on the rest the most common answer is wrong 31\% of the time. This is
the signal that self-consistency
\citep{wang2023selfconsistency}, label-free reinforcement learning \citep{Zuo2025TTRLTR} and the
concurrent u-OPSD \citep{Li2026OnPolicySW} rely on.

The model's uncertainty need not extend to nearby easier problems: a model that disagrees with
itself on a hard problem may still solve simpler related problems consistently. An old heuristic advises: if you cannot solve a problem, first solve a simpler related one
\citep{polya1945howtosolveit}. Language models can write such easier problems, called
\emph{stepping stones} (\emph{stones} for short; \citealp{Hu2026AskingTR}). On the stones Qwen3-4B writes
for itself in non-thinking mode, its mean agreement is 0.77, against 0.54 on the problems they
serve. In a separate audit of stones written and solved in thinking mode, the frontier model
Claude Opus~5 finds most
of the kept stones sound, even when the vote on the original problem is unavailable or wrong.

We introduce \ours{} (\underline{S}elf-\underline{T}aught fr\underline{O}m
\underline{N}earby \underline{E}asier problems), which keeps self-agreement but moves it from the hard problem to easier ones.
For each problem the model is unsure about, it writes a few easier related problems, solves each
several times, and keeps those it answers consistently. The teacher sees the problem together with
the worked solutions of these stones, and the student sees the problem alone. No step of \ours{}
reads a label, and the
trained model answers without stones or an extra model call.

As shown in Figure~\ref{fig:teaser}, moving agreement to easier problems produces substantially
stronger students. When Qwen~3 reasons in thinking mode, no baseline gains more than 2.2 points, while \ours{} gains 2.7 and 3.1 on Qwen3-4B and Qwen3-8B, the highest
averages. In non-thinking mode, \ours{} gains 13.2 and 22.7 points, more than twice the 5.3 and 8.5
points of OPSD with ground-truth solutions. Table~\ref{tab:main} gives the full results.
Section~\ref{sec:gemma} shows that \ours{} also matches OPSD on Gemma~4 without ground-truth
supervision. Our contributions are:
\begin{itemize}[leftmargin=*,itemsep=1pt,topsep=2pt]
  \item \textbf{Method.} Section~\ref{sec:method} introduces \ours{}, which builds label-free
  teacher context from easier problems the model writes and solves consistently, with no test-time
  overhead.
  \item \textbf{Evaluation.} Section~\ref{sec:results} compares three models in thinking and
  non-thinking settings against published and matched-recipe baselines.
  \item \textbf{Analysis.} Sections~\ref{sec:diagnosis}, \ref{sec:analysis} and~\ref{sec:audit}
  show that agreement is far stronger on stones than on the problems they serve, the kept stones
  are usually sound, and stone contexts give the largest student gains in matched-recipe
  comparisons.
\end{itemize}

\section{Background}
\label{sec:background}

\paragraph{Notation.}
We follow the notation of OPSD \citep{zhao2026self}. A training problem is $x$ and its
ground-truth solution is $y^\star$. An \emph{attempt} is a full solution
sampled from a model; its \emph{final answer} is the answer it
states at the end. Qwen~3 and Gemma~4 can work in \emph{thinking mode}, writing a long private
reasoning trace before the solution, or reason only in the visible answer, which we call
\emph{non-thinking mode}.

\paragraph{On-policy self-distillation.}
OPSD builds two policies from one language model with weights $\theta$. The \emph{student}
$p_S(\cdot \mid x)$ sees only the problem; the \emph{teacher} $p_T(\cdot \mid x, y^\star)$ also
sees the solution. The student samples an attempt $\hat y \sim p_S(\cdot \mid x)$, and both
policies score it: at each position $n$ they give probabilities for the next token,
\begin{equation*}
  p_S^{\,n} = p_S(\cdot \mid x, \hat y_{<n}), \quad p_T^{\,n} = p_T(\cdot \mid x, y^\star, \hat y_{<n}),
\end{equation*}
where $\hat y_{<n}$ is the attempt so far. Training minimizes their per-token divergence along
the student's own attempt,
\begin{equation}
  \mathcal{L}(\theta) = \frac{1}{|\hat y|}\sum_{n=1}^{|\hat y|}
  D_\delta\big(p_T^{\,n} \,\big\|\, p_S^{\,n}\big),
  \label{eq:loss}
\end{equation}
and gradients flow only through the student. The teacher never writes a solution of its own; it
only scores the student's attempt, and its extra text decides how it scores. As in the OPSD
implementation, $D_\delta$ is a term-capped forward-KL loss: each vocabulary term is capped before
the sum. Appendix~\ref{app:lr} gives the exact form. Also as
in OPSD, the teacher is the model at the start of training and stays fixed. As shown at the top of
Figure~\ref{fig:teaser}, \ours{} keeps this objective and replaces $y^\star$ by a context $z$ that
the model builds itself.
\paragraph{Agreement.}
Draw $K$ attempts at $x$ from the model. Let $\hat a(x)$ be the most common final answer, with
equivalent answers merged, and the \emph{agreement} $\alpha(x)$ the share of the $K$ attempts
that give it. We write $\tau > 1/2$ for the agreement needed to accept a vote. Attempts without
a final answer count against agreement, and ties go to the answer that appears first.

\section{Why Voting on the Problem Fails}
\label{sec:diagnosis}

\begin{figure}[t]
  \centering
  \includegraphics[width=\columnwidth]{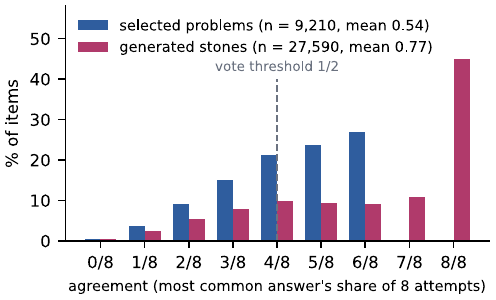}
  \caption{\textbf{Agreement on hard problems vs.\ on the stones written for them}
  (Qwen3-4B, non-thinking mode, after the basic statement filters and before applying the keep rule:
  the 9{,}210 problems it is unsure of that have at least one stone, of 9{,}218, and the 27{,}590
  stones it writes for them). Agreement is the share of 8 attempts giving the most
  common answer; the problems are selected for agreement of at most 6/8 on the same attempts, so
  their agreement is low partly by construction.}
  \label{fig:agreement}
\end{figure}

A label-free teacher could replace $y^\star$ by an attempt that reaches $\hat a(x)$, trusted
when $\alpha(x)$ is high enough, for example at least $1/2$ \citep{Li2026OnPolicySW}. The problems
worth learning from, however, are those the model is unsure about. Of the 9{,}218 problems on
which Qwen3-4B has $\alpha(x) \le 0.75$ (from 8 non-thinking attempts), 28\% have
$\alpha(x) < 1/2$ and yield no usable vote. On the other 72\%, $\hat a(x)$ is correct only 69\%
of the time, so about one in three such teachers would teach a wrong answer. Over all 29{,}434
problems the same numbers are 9\% and 87\%: votes fail mainly on the selected problems.\footnote{Ground-truth answers are
used here, and in every analysis, only to measure. Training never reads them.}
In Figure~\ref{fig:teaser}, for instance, only 3 of 8 attempts agree. The stones written for these
problems behave differently. Section~\ref{sec:method} describes how they are produced. As shown in
Figure~\ref{fig:agreement}, before the keep rule is applied, 45\% of them are answered the same way
in all 8 attempts, and their agreement averages 0.77 against 0.54 for the problems.
Section~\ref{sec:audit} checks that the answers they agree on are also mostly correct.

\paragraph{A simple illustration.}
Why are votes weak on hard problems, and why could easier problems help? Let a problem
be a chain of $k$ basic steps. The model gets step $j$ right with probability $p_j$, between
$\frac12 + \gamma$ and $\bar p < 1$, independently of the other steps. Its final answer is a
one-to-one function of the step answers, so two attempts agree only if they agree at every step.

\begin{proposition}
\label{prop:main}
(a) Two attempts at the problem give the same final answer with probability at most
$\bar p^{\,k}$, so the probability that any pair among $K$ attempts gives the same final answer is
at most $\binom{K}{2}\bar p^{\,k}$. (b) For a single step, the probability that the same wrong
answer appears in at least $\tau K$ attempts is at most
$e^{-2K(\tau-\frac12+\gamma)^2}$. If $p_j > \tau$, the probability that fewer than $\tau K$
attempts give the correct answer is at most $e^{-2K(p_j-\tau)^2}$.
\end{proposition}

Part (a) bounds the appearance of a matching pair, rather than the modal share $\alpha(x)$
directly; any vote with $\alpha(x) \ge \tau$ necessarily contains such a pair. Thus a vote on a long
problem needs exponentially many attempts before even this necessary event is likely. Part (b)
matches the threshold used by our method: a wrong answer is unlikely to reach it on a basic step,
while a step the model solves with probability above $\tau$ is likely to pass. Problems that isolate
parts of a hard one can therefore carry a signal where the problem itself carries little. This toy
model is deliberately simplified\footnote{Its assumptions exclude systematic and correlated
errors, so the bound on wrong agreement is only illustrative.} Appendix~\ref{app:theory} gives the
proof. Our claims rest on the measurements.

\section{\ours{}}
\label{sec:method}

\begin{figure*}[t]
  \centering
  \includegraphics[width=\textwidth]{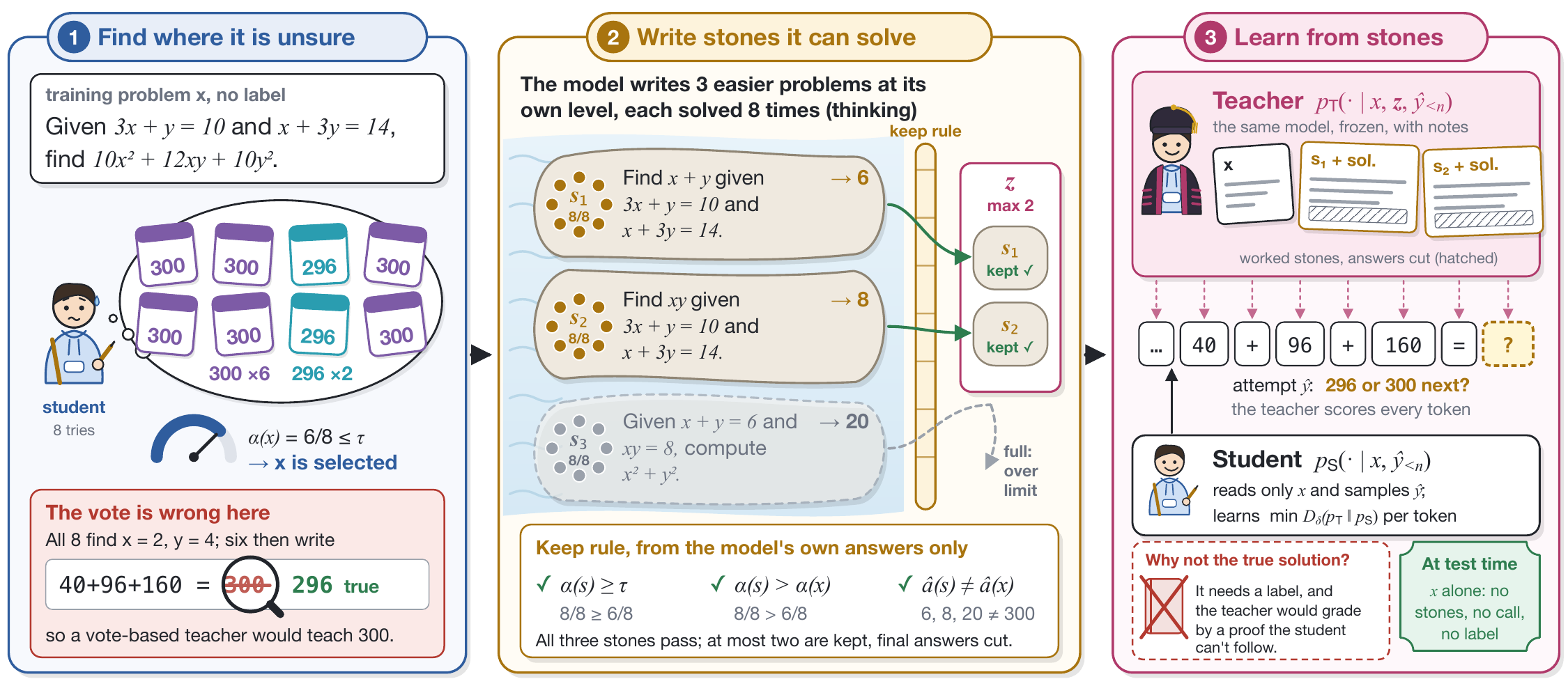}
  \caption{\textbf{The three steps of \ours{}}, on a real Qwen3-4B training problem. Step 1
  selects problems the model answers inconsistently; here 6 of 8 attempts make the same slip, so the vote on
  $x$ is wrong (the true answer, 296, is shown only for illustration). Step 2 has the model write easier related
  problems, try each 8 times, and keep those it answers consistently. This example uses the
  thinking setting. Step 3
  trains the student $p_S(\cdot \mid x)$ to match the teacher $p_T(\cdot \mid x, z)$, which also
  sees the kept problems $z$ and their worked solutions. Algorithm~\ref{alg:stone} in
  Section~\ref{sec:method} summarizes the procedure.}
  \label{fig:method}
\end{figure*}

As shown in Figure~\ref{fig:agreement}, \ours{} keeps self-agreement but moves it from the hard
problem to easier ones, where agreement is much higher. It takes training problems without
solutions and proceeds in three steps, summarized in Algorithm~\ref{alg:stone}.

\begin{algorithm}[H]
\caption{The three steps of \ours{}.}
\label{alg:stone}
\small
\begin{algorithmic}[1]
\Statex \textbf{Input:} Unlabeled problems $\mathcal{X}$, initial model $\theta_0$, and $K,M,\tau$.
\Statex \textbf{Step 1: Find uncertain problems}
\State Estimate $\hat a(x)$ and $\alpha(x)$ from $K$ attempts for each $x \in \mathcal{X}$.
\State $\mathcal{U} \gets \{x \in \mathcal{X} : \alpha(x) \le \tau\}$.
\Statex \textbf{Step 2: Pose and verify stones}
\State $\mathcal{D} \gets \emptyset$.
\For{each $x \in \mathcal{U}$}
  \State Generate $M$ candidate stones.
  \State Estimate each $\hat a(s)$ and $\alpha(s)$ from $K$ attempts.
  \State Keep up to two by Eq.~\eqref{eq:gate}, forming $z(x)$.
  \State Add $(x,z(x))$ to $\mathcal{D}$ when $z(x) \ne \emptyset$.
\EndFor
\Statex \textbf{Step 3: Distill with stones}
\State Train by Eq.~\eqref{eq:loss}, with $z(x)$ as the fixed teacher's context.
\State \Return the trained model.
\end{algorithmic}
\end{algorithm}

\paragraph{Step 1: find uncertain problems.}
Here self-agreement acts as an uncertainty signal and is measured in non-thinking mode.
Restricting stone generation to $\mathcal{U}$ concentrates it on problems whose answers are least
stable.

\paragraph{Step 2: pose and verify stones.}
Using the prompt in Appendix~\ref{app:prompts}, the model proposes easier related problems, such as
a special case, a smaller instance, or a sub-problem. Statements that repeat $x$ or another
candidate, contain a boxed answer, or fail a length check are discarded. We call each remaining
candidate a \emph{stone} $s$ and filter it with the \emph{keep rule}
\begin{equation}
  \alpha(s) \ge \tau, \quad \alpha(s) > \alpha(x), \quad \hat a(s) \ne \hat a(x).
  \label{eq:gate}
\end{equation}
In words: the model answers the stone consistently; more consistently than $x$; and with an
answer different from its most common answer to $x$. Since selected problems already have
$\alpha(x) \le \tau$, the second condition is nearly implied by the first.
Appendix~\ref{app:checks} shows that most rejected stones fail the first condition.

The third condition reflects what stones are: separate easier problems, not ordered steps toward
$x$, so none is meant to share the answer of $x$. In Figure~\ref{fig:teaser}, the stones ask for the
sum and the sum of squares of the three cosines (answers 0 and 3/2), while $x$ asks for the sum of
fifth powers. A stone that shares the answer of $x$ is usually $x$ restated, and its solution would
hand the teacher the model's vote on $x$, which is often wrong.

For each kept stone we take its shortest attempt that reaches $\hat a(s)$ and remove its final
answer. Stones are written and solved in the mode defined in Section~\ref{sec:setup}. In thinking
mode, the attempt shown is the written solution that follows the reasoning trace, without the
trace itself.

\paragraph{Step 3: distill with stones.}
For each $(x,z(x)) \in \mathcal{D}$, the context lists the kept stones and their worked solutions,
then uses the instruction in Appendix~\ref{app:prompts} to ask for an independent solution of $x$.
In Eq.~\ref{eq:loss}, the teacher is
$p_T^{\,n} = p_T(\cdot \mid x, z(x), \hat y_{<n})$, with $z(x)$ in place of $y^\star$. The student
never sees the stones, so at test time the model answers $x$ without stones or an additional model
call.

Because the teacher scores the student's attempt rather than generating a solution, stones need
not make it solve $x$ more often; Section~\ref{sec:analysis} shows that they do not.

\section{Experiments}
\label{sec:experiments}

\begin{table*}[t]
\centering
\footnotesize
\setlength{\tabcolsep}{3.2pt}
\resizebox{\textwidth}{!}{%
\begin{tabular}{lc ccccccc ccccccc}
\toprule
 & & \multicolumn{7}{c}{Thinking mode} & \multicolumn{7}{c}{Without thinking mode} \\
\cmidrule(lr){3-9} \cmidrule(lr){10-16}
Method & GT & A24 & A25 & H25 & M500 & AMC & Avg. & $\Delta$ & A24 & A25 & H25 & M500 & AMC & Avg. & $\Delta$ \\
\midrule
\multicolumn{16}{c}{\textit{Qwen3-4B}} \\
\midrule
Base &  & 74.17 & 64.72 & \underline{45.56} & 94.80 & 95.00 & 74.85 &  & 25.83 & 17.78 & 10.83 & 84.10 & 66.25 & 40.96 &  \\
\midrule
\rowcolor{gtrow}SFT & \gt & 73.06 & 69.17 & 41.67 & 95.40 & 94.38 & 74.73 & {\color{gray}-0.12} & 26.67 & 19.72 & 13.06 & 84.85 & 69.38 & 42.73 & {\color{gray}+1.77} \\
\rowcolor{gtrow}GRPO & \gt & 73.89 & \underline{69.44} & 43.61 & 95.45 & \textbf{99.38} & 76.35 & {\color{gray}+1.50} & 25.00 & 22.50 & 15.00 & 86.20 & 80.62 & 45.86 & {\color{gray}+4.90} \\
\rowcolor{gtrow}OPSD & \gt & 75.28 & 68.06 & 43.06 & 95.20 & \textbf{99.38} & 76.20 & {\color{gray}+1.35} & 32.22 & 20.83 & \underline{16.39} & 85.75 & 76.25 & 46.29 & {\color{gray}+5.33} \\
\midrule
TTRL & \nogt & 72.78 & 68.33 & \underline{45.56} & \textbf{95.90} & 96.25 & 75.76 & {\color{gray}+0.91} & 25.00 & 20.83 & 11.94 & 83.60 & 67.50 & 41.77 & {\color{gray}+0.81} \\
RENT & \nogt & 74.72 & 65.83 & 43.06 & 95.25 & \textbf{99.38} & 75.65 & {\color{gray}+0.80} & 22.22 & 20.28 & 11.39 & 84.00 & 74.38 & 42.45 & {\color{gray}+1.49} \\
Intuitor & \nogt & \underline{76.39} & 68.33 & 42.78 & 95.35 & 97.50 & 76.07 & {\color{gray}+1.22} & 23.89 & 20.00 & 11.67 & 83.70 & 70.62 & 41.98 & {\color{gray}+1.02} \\
u-OPSD\conc & \nogt & \underline{76.39} & 68.06 & \textbf{46.94} & \underline{95.75} & \underline{98.12} & \underline{77.05} & {\color{gray}+2.20} & \underline{37.50} & \underline{27.78} & 14.44 & \underline{86.50} & \underline{81.25} & \underline{49.49} & {\color{gray}+8.53} \\
\rowcolor{ourrow}\ours{} (ours) & \nogt & \textbf{77.22} & \textbf{69.72} & \textbf{46.94} & \underline{95.75} & \underline{98.12} & \textbf{77.55} & {\color{gray}+2.70} & \textbf{43.89} & \textbf{33.06} & \textbf{21.94} & \textbf{88.85} & \textbf{83.12} & \textbf{54.17} & {\color{gray}+13.21} \\
\midrule
\multicolumn{16}{c}{\textit{Qwen3-8B}} \\
\midrule
Base &  & 75.56 & 66.67 & 45.00 & \textbf{96.35} & \underline{96.88} & 76.09 &  & 27.50 & 23.33 & 13.61 & 84.05 & 69.38 & 43.57 &  \\
\midrule
\rowcolor{gtrow}SFT & \gt & 76.39 & 69.72 & 43.89 & 95.80 & 95.00 & 76.16 & {\color{gray}+0.07} & 26.94 & 21.67 & 11.94 & 84.10 & 72.50 & 43.43 & {\color{gray}-0.14} \\
\rowcolor{gtrow}GRPO & \gt & 76.94 & 69.17 & \underline{47.78} & 95.70 & 95.00 & 76.92 & {\color{gray}+0.83} & 30.56 & 21.94 & 13.06 & 87.85 & 73.75 & 45.43 & {\color{gray}+1.86} \\
\rowcolor{gtrow}OPSD & \gt & \textbf{80.83} & 69.72 & 46.67 & 95.75 & \underline{96.88} & 77.97 & {\color{gray}+1.88} & 41.67 & 28.06 & 18.33 & 87.15 & \underline{85.00} & 52.04 & {\color{gray}+8.47} \\
\midrule
TTRL & \nogt & 77.22 & 68.61 & 46.94 & 95.75 & 96.25 & 76.95 & {\color{gray}+0.86} & 27.22 & 21.11 & 13.06 & 84.45 & 71.88 & 43.54 & {\color{gray}-0.03} \\
RENT & \nogt & 77.50 & 70.28 & 45.83 & 95.95 & 96.25 & 77.16 & {\color{gray}+1.07} & 28.33 & 21.67 & 10.83 & 84.00 & 70.00 & 42.97 & {\color{gray}-0.60} \\
Intuitor & \nogt & 76.94 & 70.28 & 44.17 & \underline{96.20} & 95.62 & 76.64 & {\color{gray}+0.55} & 26.11 & 22.50 & 11.67 & 84.20 & 73.12 & 43.52 & {\color{gray}-0.05} \\
u-OPSD\conc & \nogt & 76.94 & \underline{71.39} & 47.50 & 96.00 & \textbf{98.12} & \underline{77.99} & {\color{gray}+1.90} & \underline{45.56} & \underline{34.72} & \underline{18.61} & \underline{89.55} & 83.12 & \underline{54.31} & {\color{gray}+10.74} \\
\rowcolor{ourrow}\ours{} (ours) & \nogt & \underline{79.72} & \textbf{73.61} & \textbf{48.33} & \underline{96.20} & \textbf{98.12} & \textbf{79.20} & {\color{gray}+3.11} & \textbf{65.56} & \textbf{50.00} & \textbf{32.22} & \textbf{93.65} & \textbf{90.00} & \textbf{66.29} & {\color{gray}+22.72} \\
\bottomrule
\end{tabular}}
\caption{\textbf{Without any label, \ours{} gives the highest average in both modes.} Accuracy (\%) of Qwen~3. The mode
is that of the teacher and the evaluation; the student always trains in non-thinking mode.
A24, A25, H25: AIME24, AIME25, HMMT25 (12 attempts per problem); M500, AMC: MATH500, AMC23 (4 attempts). GT: \gt{}
uses ground truth; \conc{} concurrent work. Baselines are from \citet{Li2026OnPolicySW}. $\Delta$: gain over the base
model. Best in \textbf{bold}, second \underline{underlined}, within each mode.}
\label{tab:main}
\end{table*}

\subsection{Setup}
\label{sec:setup}

\paragraph{Models and data.}
We train Qwen3-4B and Qwen3-8B \citep{Yang2025Qwen3TR} and Gemma~4 E4B \citep{Abd2026Gemma4T} (instruction
versions) on the 29{,}434 math problems of the OPSD training set \citep{zhao2026self}, which come
from OpenThoughts \citep{guha2026openthoughts}, using the problem statements only. Each model
writes its own stones. Table~\ref{tab:funnel} in Appendix~\ref{app:data} gives the counts.

\paragraph{Two settings.}
As in OPSD and u-OPSD, the student always answers in non-thinking mode during training. The
settings differ in the mode used to write and solve stones, run the teacher, and evaluate the
trained model. In the \emph{thinking} setting, all three use thinking mode; in the
\emph{non-thinking} setting, all three run without it. Thinking mode is more accurate but far more
costly: on AIME and HMMT, Qwen3-4B writes 15--19
thousand tokens per answer with it and 3--4 thousand without it, so the cheaper mode matters in
practice. Training a student without thinking mode against a teacher that reasons at length
follows OPSD, which reports that it improves the model's accuracy in thinking mode
\citep{zhao2026self}.

\paragraph{Evaluation.}
We follow the evaluation of u-OPSD \citep{Li2026OnPolicySW} exactly, so that their numbers can be compared with ours. Accuracy
is averaged over 12 attempts per problem on AIME24, AIME25 and HMMT25 (30 competition problems
each), and over 4 attempts on MATH500 \mbox{\citep{hendrycks2021measuring,lightman2024lets}} and AMC23
(500 and 40 problems), sampled at temperature 1.0 with up to 38{,}912 new tokens. Checkpoints
are saved every 25 steps and, following OPSD and u-OPSD, the best one is reported for each
benchmark.

\paragraph{Training.}
We use $K = 8$ attempts, $M = 3$ stones and $\tau = 0.75$. Only small low-rank adapter weights
are trained (LoRA, rank 64; \citealp{hu2022lora}), so the fixed teacher is the same network with
the adapter switched off. We follow the u-OPSD schedule of 150 steps of 32 problems. Relative to
the original OPSD recipe, the main change is increasing student attempts from 1{,}024 to 4{,}096
tokens. At 1{,}024 tokens, 45--68\% of the attempts during OPSD training are
cut off before they reach an answer, against 4--15\% at 4{,}096 on Qwen~3. All hyperparameters are
listed in Appendix~\ref{app:lr}.

\paragraph{Baselines.}
\emph{Main table.} For Qwen~3, the base models and all baselines in Table~\ref{tab:main} are the
numbers published by u-OPSD \citep{Li2026OnPolicySW} under the same evaluation, each with its own recipe.
Three use ground truth: supervised fine-tuning on the solutions (SFT), reinforcement learning with
answer checking (GRPO; \citealp{Shao2024DeepSeekMathPT}), and OPSD with its original 1{,}024-token
attempts. Four do not: TTRL rewards agreement \citep{Zuo2025TTRLTR}, RENT and Intuitor reward
confidence \citep{prabhudesai2026maximizing,zhao2026learning}, and u-OPSD distills from a majority-vote
attempt \citep{Li2026OnPolicySW}. \emph{Matched-recipe comparisons.} In the ablations and on Gemma~4, our
own OPSD runs use the same training recipe as \ours{}, including 4{,}096-token student attempts; with its
original recipe, OPSD in our setup closely reproduces the published numbers.
Appendix~\ref{app:opsdorig} gives the comparison. \emph{u-OPSD on Gemma~4.} With no published
numbers, u-OPSD is our reimplementation (its code is not released) at its published settings.
Appendix~\ref{app:uopsd} shows that it scores close to the published numbers on Qwen~3.

\subsection{Main Results}
\label{sec:results}

\paragraph{Thinking mode.}
The left half of Table~\ref{tab:main} reports the thinking-mode results. Here Qwen~3 is most
accurate, so long reasoning leaves little room: no published baseline gains more than 2.2 points
over the base model. \ours{} raises the average from $74.85 \rightarrow 77.55$ on Qwen3-4B (+2.7)
and $76.09 \rightarrow 79.20$ on Qwen3-8B (+3.1), the highest at both sizes, ahead of u-OPSD
(77.05 and 77.99) and of OPSD with ground-truth solutions (76.20 and 77.97), without reading any
label. The largest gains are on AIME25, by 5.0 points on Qwen3-4B and 6.9 on Qwen3-8B.

\paragraph{Non-thinking mode.}
The right half of Table~\ref{tab:main} reports results when no step uses thinking mode. \ours{}
raises the average from $40.96 \rightarrow 54.17$ on Qwen3-4B (+13.2) and
$43.57 \rightarrow 66.29$ on Qwen3-8B (+22.7), the best result at both sizes. OPSD reads a
ground-truth solution for every problem and gains 5.3 and 8.5 points, while \ours{} reads none and
gains more than twice as much. Rewarding agreement or confidence on the problem itself (TTRL,
RENT, Intuitor) adds at most 1.5 points, consistent with the weak votes of
Section~\ref{sec:diagnosis}. u-OPSD, which votes on the problem, gains 8.5 and 10.7; \ours{}
exceeds it by 4.7 and 12.0 points, with the largest margins on AIME and HMMT, where the model's
attempts agree least.

\subsection{Another Model Family}
\label{sec:gemma}

\begin{figure}[t]
  \centering
  \includegraphics[width=\columnwidth]{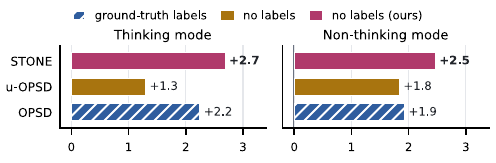}
  \caption{\textbf{Without ground-truth supervision, \ours{} matches OPSD on Gemma~4 E4B.} Gain in
  average accuracy (points) over the base model. Table~\ref{tab:gemma} gives the full results.}
  \label{fig:gemma}
\end{figure}

\paragraph{The result carries over to Gemma~4.}
Figure~\ref{fig:gemma} shows that on Gemma~4 E4B, \ours{} matches OPSD in both modes without
ground-truth supervision and is ahead of u-OPSD, though the gains over the base model are smaller
than on Qwen~3. Here OPSD uses exactly the recipe of \ours{}.

\subsection{Comparing Teacher Contexts}
\label{sec:analysis}

\begin{figure}[t]
  \centering
  \includegraphics[width=\columnwidth]{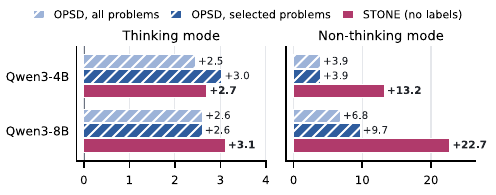}\\[3pt]
  \includegraphics[width=\columnwidth]{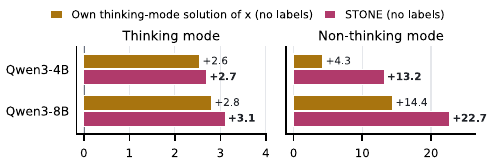}
  \caption{\textbf{In matched-recipe comparisons, stone contexts give larger student gains than a
  solution of $x$ in non-thinking mode.} Student gain over
  the base model (points), same recipe. \emph{Top:} OPSD with a ground-truth solution, on all or only
  Step~1's problems, vs.\ \ours{}. \emph{Bottom:} the model's own solution of $x$ vs.\ \ours{}. Full
  numbers in Table~\ref{tab:ablationfull}.}
  \label{fig:ablation}
\end{figure}

\begin{figure}[t]
  \centering
  \includegraphics[width=\columnwidth]{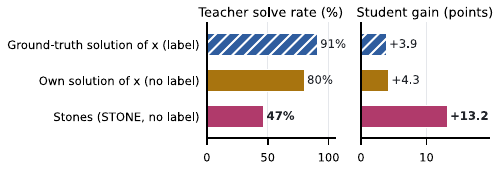}
  \caption{\textbf{Teacher solve rate does not track student gain across these contexts}
  (Qwen3-4B, non-thinking mode, same recipe). Left: teacher solve rate on the 804 problems in a
  fixed 1{,}000-problem diagnostic sample for which every diagnostic context produced complete
  outputs. Right: student gain.}
  \label{fig:teacherstudent}
\end{figure}

\paragraph{Controlled comparisons.}
We analyze non-thinking mode, where the gap to OPSD is largest, and compare teacher contexts with
the same recipe, all trained on problems Step~1 selects that have kept stones. As shown in
Figure~\ref{fig:ablation}, training with a ground-truth solution of
$x$ instead of stones lowers the student's average accuracy by 9.3 and 13.0 points on Qwen3-4B and
Qwen3-8B, and training with the model's own solution of $x$ lowers it by 8.9 and 8.3 points. This
control is also label-free: the context is the shortest of the model's 8 thinking-mode attempts at
$x$ that reaches the most common answer among them, with the final answer removed. In thinking
mode, the three contexts are within about half a point.

\paragraph{Teacher solve rate.}
Across these contexts, teacher solve rate does not track student gain. A solution of $x$ produces a
smaller gain even though it makes the teacher a far better solver. As shown in
Figure~\ref{fig:teacherstudent}, given either solution, the teacher solves most problems itself,
while with stones it solves barely more than the 46\% it solves with no context. This diagnostic starts from a fixed sample of
1{,}000 problems selected by Step~1; 804 have complete outputs for every context and form the common
comparison set.

\paragraph{Interpretation.}
A solution of $x$ can make the teacher better at solving it while tying its token scores to a path
the student cannot see. Stones instead provide worked examples that can help it score the student's own steps.
Prior work provides two relevant pieces of evidence: self-distillation can transfer reasoning
behavior rather than privileged answers \citep{Adhinarta2026DoesSI}, and worked examples can serve
as effective teacher contexts \citep{snell2023learning,Shenfeld2026SelfDistillationEC}.
Appendix~\ref{app:checks} shows that \ours{}'s checkpoints remain the most accurate at comparable
answer lengths.

\subsection{Checking the Stones}
\label{sec:audit}

\begin{table}[t]
\centering
\small
\setlength{\tabcolsep}{3.5pt}
\begin{tabular}{lcc ccc}
\toprule
 & \multicolumn{2}{c}{All stones} & \multicolumn{3}{c}{Kept, by vote on problem} \\
\cmidrule(lr){2-3} \cmidrule(lr){4-6}
 & Kept & Rejected & No vote & Wrong & Correct \\
\midrule
Stones ($n$) & 13,284 & 10,686 & 3,566 & 3,113 & 6,605 \\
\midrule
Well-posed & 86.8 & 36.6 & 85.4 & 89.3 & 86.4 \\
Correct & 90.2 & 72.6 & 88.9 & 90.5 & 90.7 \\
Sound & 78.3 & 26.6 & 75.9 & 80.9 & 78.3 \\
\midrule
Easier & 95.4 & 62.7 & 95.1 & 94.9 & 95.9 \\
Related & 96.2 & 56.5 & 95.3 & 96.3 & 96.6 \\
Restates & 4.6 & 9.0 & 4.9 & 5.1 & 4.2 \\
\bottomrule
\end{tabular}
\caption{\textbf{Most stones kept for training are sound, also where the vote on their problem fails.}
Share (\%) of Qwen3-4B's thinking-mode stones with each audited property; the last three columns split the
kept stones by the vote on their problem. ``Correct'' is among well-posed stones, and ``Sound'' means
well-posed and correct. The other 3{,}569 stones pass the keep rule but exceed the limit of two per problem.}
\label{tab:audit}
\end{table}

Agreement chooses the stones, but agreement is not correctness. We therefore have Claude Opus~5
audit all 27{,}563 stones that Qwen3-4B writes in thinking mode, using the prompt in
Appendix~\ref{app:prompts}. The audit itself uses no ground truth. The judge solves each stone and
assesses whether it is well-posed and has a unique final answer; whether the model's most common
answer and solution are correct; whether the stone is easier than the original problem and useful
for solving it; and whether it merely restates that problem. We call a stone \emph{sound} if it is
well-posed and correctly answered; 27{,}539 of the 27{,}563 replies give readable verdicts. As a
capability check, we ask the judge to solve the 5{,}757 Step~1 problems that have numeric
ground-truth answers. It answers 93.5\% correctly, indicating sufficient mathematical capability
to audit the easier stones.

\paragraph{The keep rule selects sound stones.} Table~\ref{tab:audit} shows that 78\% of the
stones kept for training are sound, against 27\% of those the rule rejects. Nearly all of them are
easier than their problem and related to it (95\% and 96\%), only 5\% restate it, and 84\% of the
solutions shown to the teacher are valid.

\paragraph{The stones hold up where the vote fails.} Section~\ref{sec:diagnosis} shows that only
half of the selected problems have a correct vote of their own. The stones written for the other
half are just as sound: 76\% when the problem has no usable vote and 81\% when its vote is wrong,
against 78\% when it is right. Agreement on easier problems thus gives the teacher a mostly correct
context where agreement on the problem gives a missing or wrong one. As Table~\ref{tab:main}
shows, the remaining flaws, about one stone in five, do not stop \ours{} from outperforming OPSD
with ground-truth solutions. Together with the matched-recipe comparisons in
Section~\ref{sec:analysis}, these results suggest that the teacher uses stones as worked examples
rather than as answers to copy.

\subsection{Choosing Problems}
\label{sec:design}

Step~1 keeps only the problems the model is unsure about. Training OPSD on these problems instead of
on all of them, with the same recipe, gives about as large a gain or a larger one in every case, as
shown at the top of Figure~\ref{fig:ablation}. The difference is clearest on Qwen3-8B in
non-thinking mode, +9.7 against +6.8 points, and on Qwen3-4B in thinking mode, +3.0 against +2.5.
Step~1 also leaves only 31\% and
30\% of the training problems to write stones for on Qwen3-4B and Qwen3-8B, which cuts the cost of
Step~2 by about two thirds.

\section{Related Work}
\label{sec:related}

\paragraph{Self-distillation.}
Distillation trains a student on a teacher's word probabilities; on-policy distillation does so
on the student's own samples \citep{agarwal2024onpolicy,gu2024minillm}. Context distillation uses one
model as both, the teacher being given extra context such as instructions or worked examples
that the student must learn to do without \citep{snell2023learning}. SDFT applies this to learning
from demonstrations \citep{Shenfeld2026SelfDistillationEC}, and OPSD to reasoning, with the teacher given a
ground-truth solution \citep{zhao2026self}. Follow-up work makes the teacher less sensitive to
that solution \citep{Saadi2026OneTM}, adapts the loss to whether the student was right
\citep{Wang2026OutcomeGuidedOS}, or distills only wrong attempts \citep{Zhao2026NotER}. All of them read ground
truth; \ours{} keeps the OPSD training objective and changes only what the teacher reads.

\paragraph{Learning without labels.}
Without answers, a model can be rewarded for agreeing with itself \citep{Zuo2025TTRLTR}, for its
confidence \citep{prabhudesai2026maximizing,zhao2026learning}, or for low uncertainty
\citep{Zhang2025RightQI}; it can also judge its own outputs \citep{Yuan2024SelfRewardingLM} or keep
its own correct reasoning \citep{zelikman2022star}. The concurrent u-OPSD shows the teacher a
majority-vote attempt \citep{Li2026OnPolicySW}. These methods rely on the model's judgment of the
problem being learned. \ours{} relies on its judgment of easier problems, where that judgment is
more consistent.

\paragraph{Easier problems as a tool.}
Breaking a problem into simpler ones is a classical strategy \citep{polya1945howtosolveit}.
Least-to-most prompting solves sub-problems in order at test time \citep{zhou2023leasttomost},
LADDER trains on simplified variants with answer checking \citep{Simonds2025LADDERSL}, and ARQ
shows that generated stepping stones help a solver at test time, choosing the useful ones with
ground-truth answers \citep{Hu2026AskingTR}. Closest to our teacher's context, analogical prompting
has the model recall related problems with solutions as in-context exemplars at test time
\citep{yasunaga2024large}; step-back prompting first asks a more abstract question
\citep{zheng2024take}, thought propagation solves analogous problems and reuses their
solutions \citep{Yu2023ThoughtPA}, and decomposed prompting delegates sub-tasks
\citep{Khot2022DecomposedPA}. \ours{} uses such self-generated problems only during training, as the
teacher's extra text, chooses them by the model's own agreement, and removes them at test time.
Related training-time ideas fine-tune on self-consistent answers \citep{huang-etal-2023-large}, distill
reasoning into direct answers \citep{Yu2024DistillingS2}, or self-propose training problems
\citep{Zhao2025AbsoluteZR,Huang2025RZeroSR}.

\section{Conclusion}

Models need not trust themselves where they are weakest. \ours{} asks a model to write nearby easier
problems, keeps those it solves consistently, and uses their worked solutions as teacher context.
This label-free context gives the strongest Qwen~3 averages in both
modes and, without ground-truth supervision, matches OPSD on Gemma~4. Useful teacher context
therefore need not reveal the target solution directly: label-free learning can instead borrow its
signal from easier problems where the model is reliable.

\section*{Limitations}

Agreement on a stone is not correctness: a stronger model judges about one in five stones used in
training ill-posed or wrongly answered, and this audit is itself model-based rather than a human
verification. We study math problems with checkable final
answers, where agreement is defined by matching answers; other tasks need another notion of
agreement. Generating and checking stones adds preprocessing time before distillation. We evaluate
models up to 8B, and whether the gains persist at larger scales remains open.

\bibliography{refs}

\appendix
\renewcommand{\topfraction}{0.9}
\renewcommand{\dbltopfraction}{0.9}
\renewcommand{\textfraction}{0.05}
\renewcommand{\floatpagefraction}{0.85}
\renewcommand{\dblfloatpagefraction}{0.85}

\section{Prompts}
\label{app:prompts}
\paragraph{Writing stones.} The model receives the prompt below in the mode of the setting, with $M = 3$.

\begin{small}
\begin{verbatim}
You are helping a student who is stuck on a hard
math problem. Following the advice "if you cannot
solve a problem, first solve a simpler related
problem", write {M} easier problems that serve as
stepping stones toward it.

Original problem:
{problem}

Requirements for each stepping-stone problem:
- It must be strictly easier than the original.
  Good choices are: a special case (small or
  concrete values), a simplified version with
  fewer conditions, or a key sub-problem whose
  result is useful for the original.
- It must be fully self-contained and well-posed,
  with exactly one final answer (a number or a
  short expression).
- It must not simply restate the original problem.
- Do NOT solve anything. Write only the problem
  statements.

Output format, exactly:
<stone> ... </stone>  (M times)
\end{verbatim}
\end{small}

\paragraph{Solving stones and problems.} Every attempt, by the student or on a stone, uses
the prompt

\begin{small}
\begin{verbatim}
Problem: {problem}

Please reason step by step, and put your final
answer within \boxed{}.
\end{verbatim}
\end{small}

\paragraph{Removing a stone's final answer.} The solution is cut at its ``Final Answer'' heading, or
otherwise at the paragraph holding the last \verb|\boxed{}|; any \verb|\boxed{}| left is unwrapped, and a
note, ``(The final answer to this related problem is omitted.)'', marks the cut.

\paragraph{Own solution as context.} For this control, the model attempts $x$ 8 times in thinking
mode, with the sampling used for stones; the context is its shortest attempt reaching the most
common answer, final answer removed, shown in the reference-solution template of OPSD.

\paragraph{Teacher context.} The teacher receives the problem and the kept stones with their worked solutions,
and an instruction that the stones are not the original problem; its last sentence follows the OPSD
teacher template.

\begin{small}
\begin{verbatim}
Problem: {problem}

Here are some simpler related problems with
worked solutions:

=== Related Problem 1 ===
{stone 1}
=== Solution to Related Problem 1 ===
{solution 1}
...
=== End of Related Problems ===

The problems above are simpler problems related
to the original problem, with worked solutions.
They are not the original problem, and their
answers are not its answer. Use any ideas from
them that help. Now, using your own independent
reasoning, solve the original problem. Think step
by step, explore different approaches, and don't
be afraid to backtrack or reconsider if something
doesn't work out:

Now solve the original problem: {problem}

Please reason step by step, and put your final
answer within \boxed{}.
\end{verbatim}
\end{small}

\paragraph{How the teacher scores the student's attempt.} Each prompt uses the model's chat template, the
student's in non-thinking mode and the teacher's in the teacher's mode. In the thinking setting, the
teacher's prompt ends with the opening of the assistant turn and the student's attempt is appended
there, as in the OPSD implementation.

\paragraph{Auditing stones.} For Section~\ref{sec:audit}, Claude Opus~5 receives the system prompt below,
with at most 8{,}192 output tokens; the user turn gives the problem the stone was written for, the
stone, the model's most common answer to it, and one of its attempts. Table~\ref{tab:audit} uses the
27{,}539 of 27{,}563 replies that end with readable verdicts. For the check of the judge, it receives
only a problem and is asked for its final answer.

\begin{small}
\begin{verbatim}
You are an expert mathematician auditing math
problems that a small language model wrote for
itself.

The model was stuck on a hard TARGET problem. It
wrote an easier related problem (the STONE),
attempted the stone several times, and its most
common answer is given as STONE ANSWER. One of its
attempts is given as STONE SOLUTION.

Your job:
1. Solve the stone yourself, carefully and
   independently. Do not trust the given answer or
   solution.
2. Judge the stone and the model's answer.

Answer each question:
- well_posed: is the stone a well-defined math
  problem with exactly one final answer?
- stone_answer_correct: is STONE ANSWER the correct
  final answer to the stone? Use null only if the
  stone is not well-posed.
- solution_valid: is STONE SOLUTION a correct
  derivation (minor slips that do not affect the
  result are fine)?
- easier_than_target: is the stone easier to solve
  than the target?
- related_to_target: does solving the stone help
  with the target (a special case, a smaller
  instance, a sub-problem, or a key step)?
- restates_target: is the stone essentially the
  target itself, possibly reworded?

Reason as briefly as needed, then end your reply
with a single line of JSON and nothing after it:
{"reference_answer": ...,
 "well_posed": ...,
 "stone_answer_correct": ...,
 "solution_valid": ...,
 "easier_than_target": ...,
 "related_to_target": ...,
 "restates_target": ...}
\end{verbatim}
\end{small}

\section{Proof of Proposition~\ref{prop:main}}
\label{app:theory}
We use the model of Section~\ref{sec:diagnosis}: a problem is a chain of $k$ basic steps
$j_1, \dots, j_k$; the model answers step $j$ by drawing from a distribution
$\pi(\cdot \mid j)$, independently across steps, and gets it right with probability
$p_j \in [\frac12 + \gamma, \bar p]$; the final answer is a one-to-one function of the step
answers, so two attempts give the same final answer exactly when they agree on every step.

\begin{proof}[Proof of Proposition~\ref{prop:main}]
(a) Two independent attempts agree on step $j$ with probability $\sum_b \pi(b \mid j)^2$. Since
$p_j > \frac12$, the correct answer is the most likely one, so
$\sum_b \pi(b \mid j)^2 \le \max_b \pi(b \mid j) \sum_b \pi(b \mid j) = p_j \le \bar p$. The steps
are independent, so the two attempts agree on all $k$ steps with probability at most
$\bar p^{\,k}$. A union bound over the $\binom{K}{2}$ pairs of $K$ attempts gives the second
claim.

(b) Let $N$ be the number of correct answers among $K$ attempts at step $j$; it is binomial with
mean $Kp_j$. If the same wrong answer appears at least $\tau K$ times, then
$N \le (1-\tau)K$. Hoeffding's inequality and $p_j \ge \frac12+\gamma$ give
\[
\Pr\!\left(N \le (1-\tau)K\right)
\le e^{-2K(p_j+\tau-1)^2}.
\]
Since $p_j \ge \frac12+\gamma$, this is at most
$e^{-2K(\tau-\frac12+\gamma)^2}$.
If $p_j>\tau$, the same inequality gives
\[
\Pr(N<\tau K) \le e^{-2K(p_j-\tau)^2}.
\]
\end{proof}

\section{Hyperparameters}
\label{app:lr}
\paragraph{Term-capped forward-KL loss.} As in the OPSD implementation, each vocabulary term of the
forward KL is capped before the sum: at position $n$,
\[
D_\delta\big(p_T^{\,n}\|p_S^{\,n}\big)
= \sum_v \min\left(p_T^{\,n}(v)\log\frac{p_T^{\,n}(v)}{p_S^{\,n}(v)},\, \delta\right),
\]
and Eq.~\ref{eq:loss} averages $D_\delta$ over the student's tokens. Because the cap is applied
termwise, $D_\delta$ is not literally a KL divergence. The values of $\delta$, listed in
Table~\ref{tab:hparams} for each setting, are the defaults of the OPSD implementation.

Table~\ref{tab:hparams} lists the hyperparameters of \ours{}. They are shared by all models and
both settings, except where a row gives values per setting or model family.

\begin{table*}[t]
\centering
\footnotesize
\setlength{\tabcolsep}{4pt}
\begin{tabular}[t]{@{}>{\raggedright}p{0.17\textwidth}>{\raggedright\arraybackslash}p{0.29\textwidth}@{}}
\toprule
\multicolumn{2}{@{}l}{\textbf{Step 1: agreement on the problem}} \\
Attempts $K$ & 8, non-thinking mode \\
Sampling & temperature 0.7, top-$p$ 0.8, top-$k$ 20 \\
Maximum length & 4{,}096 tokens \\
Selection & $\alpha(x) \le \tau = 0.75$ \\
\midrule
\multicolumn{2}{@{}l}{\textbf{Step 2: stepping stones}} \\
Stones written $M$ & 3 per problem \\
Attempts per stone & 8 \\
Mode & as the setting's teacher: non-thinking or thinking \\
Sampling & non-thinking: as Step 1; thinking: temperature 0.6, top-$p$ 0.95, top-$k$ 20 \\
Maximum length & non-thinking: 1{,}536 tokens to write, 4{,}096 to solve; thinking: 16{,}384 tokens \\
Kept stones & Eq.~\ref{eq:gate} with $\tau = 0.75$; at most 2 per problem \\
Solution shown & shortest agreeing attempt \\
\midrule
\multicolumn{2}{@{}l}{\textbf{Evaluation}} \\
Attempts per problem & 12 (AIME24, AIME25, HMMT25), 4 (MATH500, AMC23) \\
Sampling & temperature 1.0; top-$p$ 0.95 (thinking), 0.8 (non-thinking) \\
Max.\ new tokens & 38{,}912 \\
Checkpoint & best per benchmark \\
\bottomrule
\end{tabular}\hfill
\begin{tabular}[t]{@{}>{\raggedright}p{0.17\textwidth}>{\raggedright\arraybackslash}p{0.29\textwidth}@{}}
\toprule
\multicolumn{2}{@{}l}{\textbf{Step 3: training}} \\
Trained weights & LoRA, rank 64, $\alpha = 128$, dropout 0.05, all attention and MLP projections \\
Teacher & starting model (adapter off), fixed \\
Token loss $D_\delta$ & term-capped forward KL \\
Cap $\delta$ & $10^{-6}$ (non-thinking), $0.05$ (thinking) \\
Learning rate & Qwen~3: $5{\times}10^{-6}$ (non-thinking), $3{\times}10^{-6}$ (thinking); Gemma~4: $10^{-6}$ (both) \\
Optimizer & AdamW, $\beta = (0.9, 0.999)$, no weight decay \\
Schedule & constant, no warm-up \\
Gradient clip & 0.1 \\
Steps & 150, checkpoint every 25 \\
Problems per step & 32 \\
Student attempts & up to 4{,}096 tokens; temperature 1.1, top-$p$ 0.95, top-$k$ 20 \\
Precision & bfloat16 \\
\bottomrule
\end{tabular}
\caption{Hyperparameters of \ours{}. The OPSD runs of Tables~\ref{tab:gemma} and~\ref{tab:ablationfull} use the learning rates of \ours{} without tuning. The student temperature (1.1),
the 150 steps and the 32 problems per step follow the u-OPSD protocol; each step samples one
attempt per problem, and no problem is seen twice (4{,}800 of the 7k to 15k training problems
are used).}
\label{tab:hparams}
\end{table*}

Table~\ref{tab:compute} reports the measured preprocessing and training computation.
\begin{table}[!htb]
\centering
\footnotesize
\setlength{\tabcolsep}{5pt}
\begin{tabular}{lr}
\toprule
 & H100 hours \\
\midrule
\multicolumn{2}{l}{\textit{Qwen3-4B}} \\
Step 1, both modes & 10.9 \\
Step 2, non-thinking & 9.3 \\
Step 3, non-thinking & 16.3 \\
Step 2, thinking & 127.3 \\
Step 3, thinking & 16.6 \\
\addlinespace
\multicolumn{2}{l}{\textit{Qwen3-8B}} \\
Step 1, both modes & 12.8 \\
Step 2, non-thinking & 15.3 \\
Step 3, non-thinking & 22.8 \\
Step 2, thinking & 159.3 \\
Step 3, thinking & 24.3 \\
\addlinespace
\multicolumn{2}{l}{\textit{Gemma~4 E4B}} \\
Step 1, both modes & 7.1 \\
Step 2, non-thinking & 16.6 \\
Step 3, non-thinking & 34.4 \\
Step 2, thinking & 32.1 \\
Step 3, thinking & 36.4 \\
\bottomrule
\end{tabular}
\caption{Preprocessing and training computation for \ours{}. Steps~1 and~2 are
preprocessing, and Step~1 is shared by the two modes. Each GPU job used eight NVIDIA H100 GPUs.
H100 hours sum the time of parallel jobs. Evaluation is excluded.}
\label{tab:compute}
\end{table}

\section{Full Results on Gemma~4 E4B}
\label{app:gemma}
Table~\ref{tab:gemma} gives the per-benchmark results summarized in Figure~\ref{fig:gemma}.

\begin{table*}[t]
\centering
\small
\setlength{\tabcolsep}{3.4pt}
\resizebox{\textwidth}{!}{%
\begin{tabular}{lc cccccc cccccc}
\toprule
 & & \multicolumn{6}{c}{Thinking mode} & \multicolumn{6}{c}{Without thinking mode} \\
\cmidrule(lr){3-8} \cmidrule(lr){9-14}
Method & GT & AIME24 & AIME25 & HMMT25 & MATH500 & AMC23 & Avg. & AIME24 & AIME25 & HMMT25 & MATH500 & AMC23 & Avg. \\
\midrule
\multicolumn{14}{c}{\textit{Gemma 4 E4B-it}} \\
\midrule
Base &  & 56.94 & 41.39 & 28.06 & 91.20 & 85.62 & 60.64 & 43.06 & 26.94 & 20.83 & \textbf{73.25} & \textbf{76.25} & 48.07 \\
\midrule
\rowcolor{gtrow}OPSD & \gt & \textbf{58.33} & \underline{43.33} & \underline{29.72} & \textbf{92.45} & \underline{90.62} & \underline{62.89} & \underline{48.61} & \underline{31.39} & \underline{23.89} & \underline{73.00} & 73.12 & \underline{50.00} \\
u-OPSD\conc$^{\ddagger}$ & \nogt & \textbf{58.33} & 42.50 & 28.33 & 91.15 & 89.38 & 61.94 & \textbf{50.56} & 28.61 & \textbf{26.11} & 70.50 & \underline{73.75} & 49.91 \\
\rowcolor{ourrow}\ours{} (ours) & \nogt & \underline{57.78} & \textbf{44.17} & \textbf{30.28} & \underline{91.95} & \textbf{92.50} & \textbf{63.33} & 48.33 & \textbf{32.78} & 23.61 & 71.75 & \textbf{76.25} & \textbf{50.54} \\
\bottomrule
\end{tabular}}
\caption{\textbf{Without ground-truth supervision, \ours{} matches OPSD on Gemma~4 E4B.} As
Table~\ref{tab:main}, for
Gemma~4 E4B-it, with every method run by us; OPSD uses the training recipe of \ours{} (its original
recipe is in Appendix~\ref{app:opsdorig}), and $^{\ddagger}$marks our reimplementation of u-OPSD.}
\label{tab:gemma}
\end{table*}

\section{OPSD with Its Original Recipe}
\label{app:opsdorig}
Table~\ref{tab:opsdorig} compares OPSD with its original recipe \citep{zhao2026self}, which uses
1{,}024-token attempts, 100 steps and a learning rate of $5{\times}10^{-6}$, and with the recipe of
\ours{} described in Appendix~\ref{app:lr}. Table~\ref{tab:gemma} reports the matched-recipe
results in full. With the original recipe, 62--66\% of Gemma~4 E4B's training attempts are cut off
at 1{,}024 tokens, and OPSD lowers its non-thinking accuracy.

\begin{table}[!htb]
\centering
\small
\begin{tabular}{lcccc}
\toprule
 & \multicolumn{2}{c}{Non-thinking} & \multicolumn{2}{c}{Thinking} \\
\cmidrule(lr){2-3} \cmidrule(lr){4-5}
Model & original & matched & original & matched \\
\midrule
E4B-it & 47.66 & 50.00 & 61.54 & 62.89 \\
\bottomrule
\end{tabular}
\caption{Average accuracy of OPSD on Gemma~4 E4B with its original recipe and with the recipe of \ours{}
(matched).}
\label{tab:opsdorig}
\end{table}

Table~\ref{tab:opsdorigqwen} shows that on Qwen~3, OPSD with its original recipe in our setup comes
within 0.7 points of the numbers reported by \citet{Li2026OnPolicySW} in three of four cases. On Qwen3-8B
without thinking mode our run is 2.1 points lower, so the reported number, which
Table~\ref{tab:main} uses, is the stronger baseline.

\begin{table}[!htb]
\centering
\small
\begin{tabular}{lcccc}
\toprule
 & \multicolumn{2}{c}{Non-thinking} & \multicolumn{2}{c}{Thinking} \\
\cmidrule(lr){2-3} \cmidrule(lr){4-5}
Model & reported & ours & reported & ours \\
\midrule
Qwen3-4B & 46.29 & 45.60 & 76.20 & 75.78 \\
Qwen3-8B & 52.04 & 49.95 & 77.97 & 78.65 \\
\bottomrule
\end{tabular}
\caption{OPSD with its original recipe on Qwen~3: numbers reported by \citet{Li2026OnPolicySW} and our run.}
\label{tab:opsdorigqwen}
\end{table}

\section{Our u-OPSD Reimplementation}
\label{app:uopsd}
Table~\ref{tab:uopsdrepro} shows that on Qwen~3, our reimplementation of u-OPSD differs from the
numbers reported by \citet{Li2026OnPolicySW} by $-1.0$ to $+2.6$ points, with no consistent direction.
The lack of a consistent bias supports using the same implementation and settings for the
Gemma~4 comparison. \ours{} is ahead of both the reported and our u-OPSD numbers in every case.

\begin{table}[!htb]
\centering
\small
\setlength{\tabcolsep}{4pt}
\begin{tabular}{lcccc}
\toprule
 & & \multicolumn{2}{c}{u-OPSD} & \\
\cmidrule(lr){3-4}
Qwen~3 & \ours{} & reported & ours & diff. \\
\midrule
4B, non-thinking & 54.17 & 49.49 & 48.52 & $-0.97$ \\
8B, non-thinking & 66.29 & 54.31 & 56.88 & $+2.57$ \\
4B, thinking & 77.55 & 77.05 & 76.87 & $-0.18$ \\
8B, thinking & 79.20 & 77.99 & 77.66 & $-0.33$ \\
\bottomrule
\end{tabular}
\caption{Average accuracy of our u-OPSD reimplementation on Qwen~3, against the numbers reported by
\citet{Li2026OnPolicySW} and \ours{}.}
\label{tab:uopsdrepro}
\end{table}

\section{Training Data and Teacher Solve Rates}
\label{app:data}
Table~\ref{tab:funnel} gives the training data each model builds, and Table~\ref{tab:context} the teacher
solve rates of Figure~\ref{fig:teacherstudent}.

\begin{table}[!htb]
\centering
\footnotesize
\begin{tabular}{lcc}
\toprule
Teacher context & Solve rate (\%) & $\Delta$ \\
\midrule
No context & 46.1 &  \\
\rowcolor{ourrow}Stones (\ours{}) & 47.0 & +0.9 \\
Own solution of $x$ & 80.2 & +34.1 \\
Ground-truth solution (OPSD) & 91.0 & +44.9 \\
\bottomrule
\end{tabular}
\caption{\textbf{Teacher solve rates of Figure~\ref{fig:teacherstudent}.} How often the teacher
(Qwen3-4B, non-thinking mode), given each context, solves the 804 problems in a fixed
1{,}000-problem diagnostic sample for which every diagnostic context produced complete outputs
(8 attempts each, checked against ground truth for analysis only). $\Delta$: change from no context.}
\label{tab:context}
\end{table}

\begin{table}[!htb]
\centering
\footnotesize
\setlength{\tabcolsep}{2.5pt}
\begin{tabular}{lcccc}
\toprule
 & & & \multicolumn{2}{c}{With stones} \\
\cmidrule(lr){4-5}
Model & Problems & Selected & non-thinking & thinking \\
\midrule
Qwen3-4B & 29,434 & 9,218 & 7,645 & 7,338 \\
Qwen3-8B & 29,434 & 8,706 & 7,302 & 8,136 \\
Gemma~4 E4B & 29,434 & 15,842 & 14,057 & 15,084 \\
\bottomrule
\end{tabular}
\caption{Training data built by each model. Selected: agreement $\alpha(x) \le 0.75$ (Step~1, non-thinking
attempts). With stones: at least one stone passes the keep rule (Step~2), with stones written and solved without
thinking mode (non-thinking setting) or with it (thinking setting).}
\label{tab:funnel}
\end{table}

\begin{table}[!htb]
\centering
\footnotesize\setlength{\tabcolsep}{3.5pt}
\begin{tabular}{lcccc}
\toprule
 & \multicolumn{2}{c}{Thinking} & \multicolumn{2}{c}{Non-thinking} \\
\cmidrule(lr){2-3} \cmidrule(lr){4-5}
 & 4B & 8B & 4B & 8B \\
\midrule
Base model & 74.85 & 76.09 & 40.96 & 43.57 \\
\midrule
\rowcolor{ourrow}\ours{} & 77.55 & 79.20 & 54.17 & 66.29 \\
Stones from the other mode & 77.31 & 78.83 & 51.68 & 65.46 \\
Own solution of $x$ & 77.40 & 78.90 & 45.23 & 58.01 \\
\midrule
OPSD & 77.30 & 78.70 & 44.85 & 50.36 \\
OPSD, Step~1's problems & 77.88 & 78.69 & 44.86 & 53.25 \\
\ours{}, no Step~1 & 76.68 & 78.81 & 53.78 & 67.27 \\
\bottomrule
\end{tabular}
\caption{\textbf{Full ablation grid.} Average accuracy over the five benchmarks for Qwen3-4B and Qwen3-8B. ``Other mode'' uses
thinking-mode stones in the non-thinking setting and non-thinking stones in the thinking setting. ``Own
solution of $x$'' is the model's thinking-mode solution with the final answer removed. OPSD uses the training
recipe of \ours{}, on all problems or only on those Step~1 selects; ``no Step~1'' writes stones for all
problems.}
\label{tab:ablationfull}
\end{table}

\begin{figure}[t]
  \centering
  \includegraphics[width=\columnwidth]{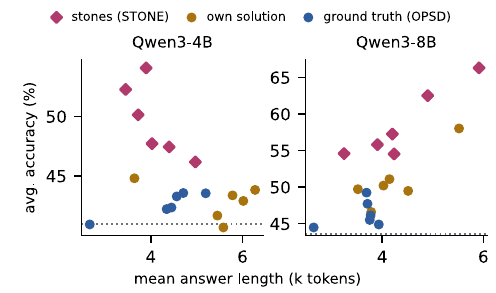}
  \caption{\textbf{At similar response lengths, \ours{} checkpoints are more accurate.} Average
  accuracy over the five benchmarks against mean answer length at every saved checkpoint
  from steps 25 to 150 for the student trained without thinking mode with three teacher contexts and
  the same recipe. The dotted line marks the base model.}
  \label{fig:dynamics}
\end{figure}

\section{Additional Checks}
\label{app:checks}
\paragraph{Keep rule.} Of the stones Qwen3-4B writes for its selected problems, the first condition of
Eq.~\ref{eq:gate} removes 35\% (non-thinking) and 31\% (thinking), the second 3\% and 1\% more, and the
third 4\% and 7\%; 58\% and 61\% pass all three.

\paragraph{Stones that reach the problem's answer.} Measured with ground truth, a kept stone's answer
equals the problem's true answer in 2.7\% (non-thinking) and 5.7\% (thinking) of the Qwen3-4B training
problems. Since kept stones differ from the model's most common answer, in these cases the stone finds an
answer that the vote on the problem missed.

\paragraph{Response style.} Across AIME24, AIME25 and HMMT25, 71\% of Qwen3-4B's non-thinking
answers after \ours{} contain ``Wait'' and 53\% contain ``Alternatively'', compared with 30\% and
6\% after matched-recipe OPSD and 35\% and 14\% before training. These markers describe a shift
toward more explicit reconsideration, not its cause: on the same problem, answers with either
marker are not more accurate than answers without one.

\paragraph{Overlap with the benchmarks.} Counting a benchmark problem as overlapping if any
13-word sequence also appears in a training problem, 8 of 30 AIME24, 10 of 30 AIME25, 0 of 30
HMMT25, 45 of 500 MATH500 and 11 of 40 AMC23 problems overlap, mostly through stock phrases. No
AIME, HMMT or AMC23 problem shares half of its 13-word sequences with any training problem; 4
MATH500 problems do, all templated exercises (for example, converting a point to polar
coordinates). All methods train on problems from the same training set.

\paragraph{Answer length.} Every teacher context makes the student's answers longer, from about 3
thousand tokens to 4 to 6 thousand. As shown in Figure~\ref{fig:dynamics}, at comparable lengths
\ours{} is the most accurate context, so its gain does not come from longer answers alone.

\end{document}